\documentclass[pdflatex,sn-mathphys]{sn-jnl}

\usepackage{natbib}
\usepackage{graphicx}%
\usepackage{multirow}%
\usepackage{amsmath,amssymb,amsfonts}%
\usepackage{amsthm}%
\usepackage{mathrsfs}%
\usepackage[title]{appendix}%
\usepackage{xcolor}%
\usepackage{textcomp}%
\usepackage{manyfoot}%
\usepackage{booktabs}%
\usepackage{algorithm}%
\usepackage{algorithmicx}%
\usepackage{algpseudocode}%
\usepackage{listings}%

\theoremstyle{thmstyleone}%
\theoremstyle{thmstyletwo}%

\theoremstyle{thmstylethree}%

\begin{document}

\title[Article Title]{Multi-View Graph Feature Propagation for Privacy Preservation and Feature Sparsity}

\author[]{\fnm{Etzion} \sur{Harari}}\email{etzionharari@mail.tau.ac.il}

\author[]{\fnm{Moshe} \sur{Unger}}\email{mosheunger@tauex.tau.ac.il}


\affil[]{\orgname{Tel Aviv University}, \orgaddress{Tel Aviv, \country{Israel}}}

\abstract{Graph Neural Networks (GNNs) have demonstrated remarkable success in node classification tasks over relational data, yet their effectiveness often depends on the availability of complete node features. In many real-world scenarios, however, feature matrices are highly sparse or contain sensitive information, leading to degraded performance and increased privacy risks. Furthermore, direct exposure of information can result in unintended data leakage, enabling adversaries to infer sensitive information.
To address these challenges, we propose a novel \textbf{Multi-view Feature Propagation (MFP)} framework that enhances node classification under feature sparsity while promoting privacy preservation. MFP extends traditional Feature Propagation (FP) by dividing the available features into multiple Gaussian-noised views, each propagating information independently through the graph topology. The aggregated representations yield expressive and robust node embeddings. This framework is \emph{novel} in two respects: it introduces a mechanism that improves robustness under extreme sparsity, and it provides a principled way to balance utility with privacy.
Extensive experiments conducted on graph datasets demonstrate that MFP outperforms state-of-the-art baselines in node classification while substantially reducing privacy leakage. Moreover, our analysis demonstrates that propagated outputs serve as \emph{alternative imputations} rather than reconstructions of the original features, preserving utility without compromising privacy. A comprehensive sensitivity analysis further confirms the stability and practical applicability of MFP across diverse scenarios. Overall, MFP provides an effective and privacy-aware framework for graph learning in domains characterized by missing or sensitive features.}


\maketitle

\section{Introduction}\label{sec1}

Graphs have become a dominant representation for modeling relational data in domains such as social networks, biology, and knowledge bases. The field of Knowledge Discovery from Graphs (KDG) has long studied methods to extract patterns and insights from such structures. Graph Neural Networks (GNNs) \citep{gnn2024} have emerged as a powerful paradigm for KDG, combining deep learning with graph structure to achieve state-of-the-art performance in diverse graph learning tasks, specifically in node classification. The GNN architecture relies heavily on rich node features to achieve strong predictive performance. However, in real-world applications, such as social networks, e-commerce platforms, and healthcare systems, these features often include \textbf{sensitive user information} (e.g., demographics, gender, preferences, or personal and behavioral attributes). Directly using such features raises concerns about \textbf{data leakage}, i.e., unintentional exposure of private data, creating serious privacy risks \citep{risk24_planetoid, risk24}. 

Recent work has begun to explore feature sparsity as a strategy for mitigating these privacy risks \citep{FP, gcnmf, risk24_planetoid}. Feature sparsity entails using a small selected subset of features for model training (rather than the full set of available attributes), thereby limiting the amount of sensitive information exposed to the learning process and decreasing the likelihood of disclosing personally identifiable information or data leakage. From a privacy perspective, this approach serves as a form of information minimization: Restricting the volume of features that are stored, processed, and shared reduces the potential attack surface for adversaries. Furthermore, a sparse representation can mitigate risks of data reconstruction or attribute inference attacks (AIA), as adversaries gain access to fewer signals that could be exploited to recover private attributes.

Yet, alongside these privacy benefits, feature sparsity poses challenges for model performance. In particular, many GNN models assume a full feature matrix and are therefore prone to performance degradation when features are absent \citep{gcnmf, FP, pagnn}. Notably, this concern is not unique to exogenously induced feature sparsity driven by privacy concerns; it also emerges in many real-world scenarios where feature information is noisy or incomplete \citep{SPIN, gcnmf, pagnn, FP}.

Feature propagation (FP) is a key approach for compensating for feature sparsity in these cases; this approach entails iteratively diffusing and sharing information across neighboring nodes, thereby compensating for missing data in a scalable manner \citep{FP}. Nevertheless, this solution has several drawbacks, also in privacy-sensitive scenarios. In particular, classical FP achieves substantially lower performance compared to models trained on a complete feature matrix \citep{FP, gcnmf}. In addition, feature sparsity in combination with FP may still be insufficient to mitigate privacy concerns, as the underlying goal of FP is to reconstruct missing features on the basis of available data. Thus, if the initial set of retained features includes sensitive attributes (e.g., in a healthcare graph where patient nodes are connected based on hospital visits, FP inputs might contain private information such as medical conditions or treatments). Moreover, there is a possibility that the propagated feature matrix might reconstruct the sensitive data that were previously obscured, negating the efforts to prevent exposure.

To address these limitations, we propose a novel framework called Multi-view Feature Propagation (MFP) for node classification tasks. The central idea of this framework is to extend classic FP by propagating knowledge from multiple, partially noised feature views rather than directly relying on original, potentially sensitive features or unavailable attributes. This principle enables MFP to improve on classic FP in two ways: (1) Privacy protection: Both FP and MFP operate on a small set of retained features; however, whereas FP consists of a single propagation step that incorporates all retained features that remain visible to downstream classification models, MFP entails multiple propagation steps, each operates incorporating an even smaller, randomly sampled subset of the retained features, which are further masked by noise. This approach limits the exposure of sensitive information while still enabling nodes to exchange useful signals with their neighbors. In this way, MFP redefines FP from a reconstruction-oriented process to one that explicitly prioritizes privacy protection. (2) Performance enhancement: Integrating multiple feature views (as opposed to a single view, as in classic FP) yields richer and more balanced node representations, reducing overfitting to any single attribute subset and improving generalization for node classification tasks.

Through this multi-view propagation mechanism, MFP improves the node classification performance of classic FP, while introducing a privacy-preserving learning paradigm tailored for graph-based tasks involving sensitive data. To the best of our knowledge, this is the first approach to jointly address sparsity and privacy in feature propagation for node classification tasks.

We conduct extensive experiments on both real-world datasets \citep{planetoid} and synthetic datasets \citep{mixhop}. Our results show that MFP consistently outperforms state-of-the-art baseline approaches, including classic FP, under high sparsity while effectively mitigating privacy risks. Moreover, we specifically investigate the potential risk of diffusing sensitive features through propagation-based methods. In addition, we perform a systematic analysis of MFP’s sensitivity to graph homophily, propagation depth, and number of views, offering \textbf{practical guidelines} for deploying MFP in diverse real-world node classification scenarios. Our full implementation of MFP is publicly available. \footnote{\url{https://github.com/EtzionR/MFP}}

In achieving a favorable trade-off between predictive accuracy and privacy protection, MFP has significant industrial implications: organizations can leverage user data for advanced analytics and recommendations without violating privacy regulations. By masking and propagating only partial views of user features, companies can safely extract insights and maintain compliance, avoiding costly legal risks and reputational damage. More generally, beyond privacy-sensitive scenarios, our approach can improve GNN node classification performance in situations of incomplete or noisy data, a common challenge in real-world applications such as e-commerce personalization, financial fraud detection, or healthcare analytics \citep{gnn2024}.

In summary, this paper makes three key contributions:
(1) it introduces MFP, a novel privacy-aware graph-learning framework that operates over multiple masked feature views;
(2) it provides comprehensive empirical evidence and sensitivity analyses demonstrating superior accuracy–privacy trade-offs under extreme sparsity; and
(3) it offers practical deployment insights for privacy-constrained domains such as personalization, finance, and healthcare.





\section{Background and Related Work}

\subsection{Graph Neural Networks for KDG}

Knowledge discovery from graph datasets plays a central role in modern data science, across domains such as social networks, biological systems, and scientific co-authorship networks \citep{gnn2024}. In the last decade, GNNs have become the dominant technique in KDG, and are widely applied to various tasks, including node classification. These models are fundamentally transforming the analysis and modeling of complex, relational data.

Numerous GNN architectures have been developed; among these, Graph Convolutional Networks (GCNs) \citep{gcn} and Graph Attention Networks (GATs) \citep{gat} are widely used in various machine learning applications, including node classification and regression. These architectures use both node attributes (features) and graph structure (edges) to learn expressive representations from attributed graphs. In these models, node features are represented by a matrix $X \in \mathbb{R}^{(|V| \times d)}$, where $V$ denotes the set of nodes and $d$ is the feature dimension; each row $X_{i \cdot}$ in $X$ is a feature vector comprising the attributes of node $i$ (for example, if the graph represents a social network, $X_{i \cdot}$ might contain user $i$’s demographic information, interests, or activity patterns). The graph topology is represented by an edge set $E$, where $(i, j) \in E$ if nodes $i$ and $j$ are connected by an edge. These two components together define the graph as $G = \{X, E\}$, where $G$ captures both the node attributes (features) and the structural relationships (edges).

GNNs leverage \textit{homophily} to learn a node’s representation from its own attributes and those of its neighbors. Homophily refers to the tendency of similar entities to form connections with one another \citep{homophily2020}. For example, in social networks, individuals might form ties with others who share comparable attributes such as age, interests, education, or political orientation. The capacity of GNNs to leverage homophily has led to remarkable success across a wide range of real-world applications, from social network analysis and recommendation systems to molecular property prediction and physical system simulation \citep{gnn2024, gnnapp2020}.



\subsection{Feature Sparsity for Privacy Preservation}
\label{subsec:gnn_sparse}

In many real-world cases, the node features and attributes represented in the feature matrix $X$ encode sensitive personal information; for example, in a social network, such features might include demographic, behavioral, or private details such as age, income, health status, or political affiliation \citep{gnn2024_privacy, social_networks23_privacy}. If exposed, such data can directly compromise individual privacy—for example, revealing a person’s medical condition, disclosing sexual preference, or exposing political orientation without consent \citep{risk24, risk24_planetoid}. In social networks, this could mean that details individuals intentionally keep private from other users become visible to adversaries. These risks highlight the necessity of privacy-preserving mechanisms that protect node-level feature data in datasets such as social networks from unauthorized exposure.

In response to this privacy risk, a body of work on privacy-preserving graph learning has emerged. One line of defense is applying differential privacy (DP) to GNNs \citep{random_dp_gnn}: for instance, adding noise to graph embeddings, aggregate messages, or training gradients so that the influence of any single node or edge is bounded. A complementary strategy is graph anonymization, i.e., sanitizing or perturbing the graph data before learning. Though these strategies can obscure sensitive details, they often come at the cost of performance, where the learned embeddings from an overly anonymized graph may be much less informative and reduce the overall prediction performance compared with reliance on the full set of features. 

Another widely discussed strategy for enhancing privacy in graph datasets is to increase the sparsity of the feature matrix $X$ \citep{feature_sparsity}. The underlying intuition is straightforward: by omitting, masking, or removing sensitive attributes from $X$, the risk of disclosing private details about individuals is reduced, since adversaries have fewer direct signals from which to infer sensitive information. For example, demographic or behavioral attributes that might expose health status, income level, or political orientation can be deliberately excluded to mitigate privacy concerns.

However, this approach introduces a fundamental trade-off between privacy and utility. As the sparsity of $X$ increases, the amount of information available to downstream models is reduced, weakening their ability to capture node characteristics and relational patterns. This degradation often manifests in poorer predictive accuracy, weaker generalization, and limited capacity to exploit the rich contextual information present in the original feature set. In many real-world scenarios, where feature sparsity may already exist due to missing data or incomplete records, deliberately enforcing further sparsity to protect privacy can exacerbate these challenges, leaving models with insufficient signal to perform effectively. In other words, this situation creates a fundamental trade-off: stronger privacy typically leads to lower predictive performance, requiring a proper solution to balance between data protection and model utility \citep{trade_off1, trade_off2, trade_off3}.

\subsection{Feature Propagation}\label{preliminary.fp}

Feature Propagation (FP) is a technique for imputing or compensating for missing node features in graph-based machine learning \citep{FP}. FP is a useful tool in efforts to address the trade-off outlined in Section \ref{subsec:gnn_sparse}. The approach is applied in settings in which the node feature matrix $X$ is sparse, with the majority of entries unobserved, and where only a limited subset $k$ of features is available. FP compensates for the missing features by leveraging the graph topology to iteratively diffuse available feature information across neighboring nodes. Through this process, nodes with incomplete or missing features can enrich their representations by aggregating information from structurally related nodes. In effect, FP mitigates the impact of sparsity on model performance: even when direct attribute information is unavailable, the relational context enables nodes to approximate missing features and generate meaningful embeddings for node classification. This property makes FP a natural candidate for addressing both sparsity-induced performance degradation and the broader challenge of learning in incomplete or noisy graph environments.

The most important condition for effective FP is homophily \citep{FP}, a principle that, as noted above, also underlies most GNN models, including GCN \citep{gcn} and GAT \citep{gat}. FP leverages the similarity between neighboring nodes in the graph to propagate feature information and compensate for missing values based on the given topology.

In practice, FP iteratively propagates the feature matrix $X$ using the normalized adjacency matrix $\hat{A}$. This matrix is derived from the classic adjacency matrix $A \in \mathbb{R}^{(|V| \times |V|)}$, where $A_{ij} = 1$ if $(i,j) \in E$ and $0$ otherwise. The normalization of $A$ is computed as follows: $\hat{A}_{ij} = \frac{1}{\sqrt{|\mathcal{N}_i| \cdot |\mathcal{N}_j|}}$, where $\mathcal{N}_i$ denotes the set of neighbors of node $i$ (for cases where $(i,j) \notin E$, then $\hat{A}_{ij} = 0$).

FP uses $\hat{A}$ to diffuse features across the graph topology. The main procedure used to do so is message passing \citep{gnn2005, mp2018}: For each iteration $\iota$, FP multiplies $X$ by $\hat{A}$ to diffuse the existing features across the graph: $H^{(\iota)}=\hat{A}H^{(\iota-1)}$ where $H^{(0)}=X$. After each multiplication, the known features are reset to their original values: $H^{(\iota)}_k=X_k$. The propagation process repeats $\gamma$ times, until convergence. Ultimately, we obtain $\hat{X}$, the output matrix of the propagation process ($\hat{X}=H^{(\gamma)}$).

The FP mechanism builds upon the concept of Dirichlet Energy (DE), a principle that has been employed in various graph learning tasks \citep{de1, de2, de3}. DE quantifies the smoothness or variability of a function across its domain. Formally, for a function $f: \Omega \rightarrow \mathbb{R}$, DE is defined as $DE(f) = \frac{1}{2} \int_{\Omega} || \nabla f(x) ||^2 dx$. DE measures the total squared magnitude of the $\nabla f$ over the domain. In a graph context, minimizing DE encourages smoothness across the graph Laplacian matrix. When node attributes are missing, FP leverages the DE principle by treating the existing attributes as fixed boundary conditions and iteratively diffuses information over the graph structure to impute unknown values. The diffusion process is designed to minimize the DE on the graph, ensuring that the resulting node features vary minimally across neighboring nodes, thereby producing smooth values that aligned with $G$ topology. Additionally, resetting the known feature set $k$ at each iteration helps prevent diverges of the resulted matrix values.

However, naive FP still suffers from two major limitations. First, models trained on $\hat{X}$ generally achieve lower performance compared to those trained on the full feature matrix $X$. Second, the subset of features $k$ used for FP, which may still contain sensitive information, is clearly observable to downstream models, thereby posing potential privacy risks. Moreover, the propagated matrix $\hat{X}$ can also be regarded as a reconstructed version of the original feature matrix $X$, which may pose more serious privacy risks. Such reconstructed representations can inadvertently retain or approximate sensitive information, making it possible for adversaries to infer private details.

In the following sections, we propose a graph learning approach that directly addresses these limitations of FP, simultaneously mitigating privacy risks and enhancing predictive performance. Moreover, we would be carefully examined the risk of reconstructed sensitive features using FP, to ensure that privacy preservation is not compromised.


\section{Method}
In this section, we present the Multi-View Feature Propagation (MFP) framework for node representation learning with privacy preservation. 
Formally, the input is an attributed graph $G=\{X,E\}$, where $E$ is the set of edges, and $X \in \mathbb{R}^{|V|\times d}$ the node feature matrix containing potentially sensitive attributes (where $V$ is the set of nodes and $d$ is the feature dimension). Our target task is to predict the node classification $\hat{Y} \in \mathbb{R}^{|V|}$ by using both the node features and the graph topology. To achieve high prediction performance while preventing exposure of the sensitive attributes in $X$, MFP constructs an alternative feature representation $\stackrel{*}{X}$ that reduces reliance on raw sensitive features while maintaining predictive power.

\begin{figure}[ht]
    \centering
    \includegraphics[width=.99\textwidth]{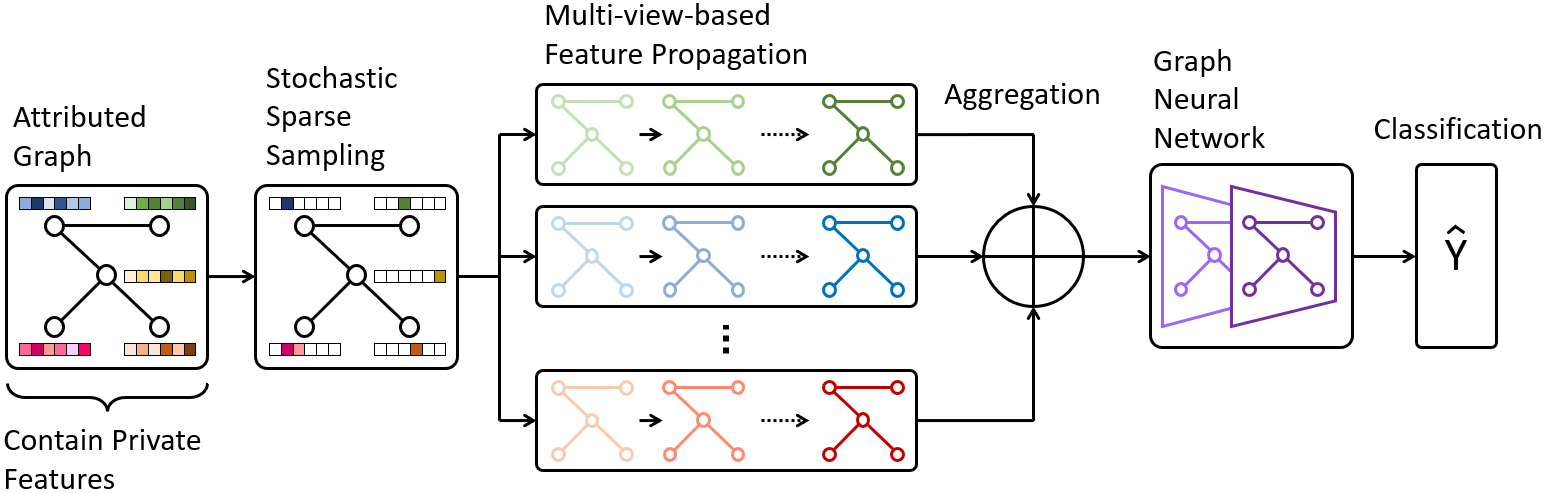}
    \caption{Multi-View Feature Propagation framework for Privacy Preservation}
    \label{pipeline}
\end{figure}

As illustrated in Figure~\ref{pipeline}, MFP follows a sequential pipeline: 

\begin{enumerate}
    \item Stochastic sparse sampling replaces most of the original features in $X$ with noise while retaining a small random subset $k$, thereby obfuscating sensitive attributes (see Section~\ref{algo.sss}).

    \item Multi-view-based Propagation then generates $\eta$ complementary propagated views by diffusing masked subsets of $k$ across the graph structure, as further detailed in Section~\ref{algo.mvfp}.

    \item The propagated views are aggregated through concatenation into a rich representation $\stackrel{*}{X}$ (Equation~\ref{concat}), which is provided to a downstream GNN for final classification.
\end{enumerate}

This design enables MFP to maintain high predictive accuracy under conditions of extreme feature sparsity, while substantially mitigating the risk of sensitive attribute leakage.

\subsection{Stochastic Sparse Sampling}\label{algo.sss}

The first component of the MFP pipeline is stochastic sparse sampling, which introduces controlled randomness into the node feature matrix before any propagation occurs. The central goal is to obscure sensitive attributes while retaining enough signal to support downstream learning tasks. We apply a stochastic sampling function $f(\cdot)$ to the node feature matrix $X \in \mathbb{R}^{|V|\times d}$ to mitigate privacy risks. Instead of exposing the full feature set, we retain only a small random subset $k$ of node features in $X$, while all remaining entries are replaced with Gaussian noise, as shown in Equation~\ref{SMFP}. This randomization procedure, which obfuscates the majority of the original attributes, is inspired by techniques in differential privacy \citep{random_dp, random_dp_2, random_dp_gnn}, where controlled noise injection makes it statistically infeasible for adversaries to reconstruct sensitive information. The resulting perturbed matrix $\stackrel{\sim}{X}$ thus serves as a privacy-preserving surrogate of $X$, suitable for subsequent propagation.

Formally, for each node $V_i$ in the graph, only a fraction of the original features are randomly sampled from the node’s feature vector $X_{i \cdot} \in \mathbb{R}^{d}$. A feature value $X_{ic}$ (where $c$ is column index in $X$) is sampled only if $X_{ic} \in k$, while the remaining values are replaced with Gaussian noise $\epsilon_{ic}$. The noise is sampled as: $\epsilon_{ic} \sim \mathcal{N}(\mu, \sigma^2)$, where $\mu$ and $\sigma^2$ denote the mean and variance, respectively. This stochastic sampling process is summarized in Equation \ref{SMFP}.

\begin{equation}\label{SMFP}
\stackrel{\sim}{X}_{ic} =
\begin{cases}
  X_{ic}, & \text{if } X_{ic} \in k \\
  \epsilon_{ic}, & \text{if } X_{ic} \not \in k
\end{cases}
\end{equation}

The output $\stackrel{\sim}{X}$ effectively hides most sensitive attributes; however, on its own, it is not expressive enough for accurate node prediction. In the next section, we describe how MFP transforms $\stackrel{\sim}{X}$ into a richer representation $\stackrel{*}{X}$ through multi-view feature propagation, enabling high performance node classification while maintaining strong privacy guarantees.

\subsection{Multi-view Feature Propagation}\label{algo.mvfp}

The next component of the framework is multi-view learning for feature propagation, which enriches the surrogate representation $\stackrel{\sim}{X}$ through the generation of multiple complementary propagated feature views. Each of these complementary views is based on a randomly sampled subset of $k$. The motivation is that no single masked subset of attributes is sufficient to construct meaningful node embeddings, but combining several partial, independently diffused views can recover the predictive signal while limiting privacy leakage. 

This approach adapts the general idea of multi-view learning \citep{mv_survey,mv_survey2}, where each data point is represented from several distinct perspectives, or views. Each view may emphasize different subsets of features, relationships, or modalities, and the integration of these perspectives yields a representation that is richer, more robust to noise, and less biased toward any single feature source.

In the context of graphs, the multi-view principle is particularly powerful. A single masked subset of node attributes may be too sparse or noisy to support effective learning. By creating multiple $\eta$ views through stochastic sampling of $k$ followed by feature propagation, MFP captures diverse patterns of attribute and structural information. Each view diffuses a different $k_t \subseteq k$ slice of the retained features across the graph topology. Each slice $k_t$ is sampled with parameter $p \in [0,1]$ from $k$, where different slices might overlap with one another, but always maintain Equation \ref{k_limit}:

\begin{equation}\label{k_limit}
|\bigcup_{t=1}^{\eta} k_t | = | k_1 \cup k_2 \cup \cdot\cdot\cdot \cup k_\eta| \leq |k|
\end{equation}

 This process ensures that only $|k|$ features from $X$ are available to the model (same number of features as in naive FP), which guarantees feature sparsity. However, the multiple views also produce robust embeddings that reflect both the local neighborhood structure and the sampled attributes. When these views are later aggregated, the resulting representation $\stackrel{*}{X}$ integrates complementary signals that would be inaccessible from any single view alone. Intuitively, each view propagates a different random projection of the feature space, thereby emphasizing distinct node attribute across the graph. Although each individual propagation step operates on a noisy or incomplete subset of features, the ensemble of views collectively captures a broader set of structural dependencies. This mechanism is analogous to ensemble averaging or dropout regularization: noise patterns introduced in individual views tend to cancel out, while stable correlations and consistent neighborhood signals are reinforced. As a result, the aggregated representation $\stackrel{*}{X}$ yield high performance for node classification tasks, while preserving sparsity and improving robustness to missing or noisy attributes.

Recall that the naive FP method relies on a single propagation step, in which $\hat{X}$ is an input matrix, where each $X_{ic}=0$ if $X_{ic} \notin k$. In our method, we initiate a separate propagation step for each subset $k_t \subseteq k$; for each of these steps, we create an input matrix $\stackrel{\sim}{X}^{(t)}$ in which elements not in $k_t$ are replaced with Gaussian noise. While these noise-imputed features may not appear particularly effective for the learning task, their propagation across the graph can construct meaningful representations, as shown in \cite{RFP}.
It is also important to note that employing a single view ($\eta = 1$), in which all features from $k$ are sampled (where $p=1$), is equivalent to applying naive FP.

\subsection{MFP Framework}\label{algo.framework}
The complete MFP pipeline is summarized in Algorithm~\ref{algo}. The framework receives as input a graph $G = \{X,E\}$, the number of required views $\eta$, the view sampling ratio $p \in [0,1]$, the number of propagation iterations ($\gamma$) and the set of retained features $k$. Its output is a prediction $\hat{Y}$ obtained through a downstream GNN model. The process unfolds as follows:

First, we initialize the surrogate matrix $\stackrel{\sim}{X}$ by the stochastic sparse sampling function $f(\cdot)$ (Section~\ref{algo.sss}). Subsequently, for each view $t \in \{1, \dots, \eta\}$, the following steps are performed:

\begin{enumerate}

    \item A distinct subset $k_t$ of features is randomly sampled from the retained set $k$ in line \ref{algo.state.get_kt} in Algorithm \ref{algo} according to $p$, with $k_t \subseteq k$. This sampling strategy ensures that each view, across all iterations, provides a different perspective on the attributes contained in $k$.

    \item Next, a noised feature matrix $\stackrel{\sim}{X}^{(t)}$ is constructed in line \ref{algo.state.get_Xt} in Algorithm \ref{algo}, containing only values from $k_t$ while all other entries are replaced with Gaussian noise.

    \item Using $\stackrel{\sim}{X}^{(t)}$, FP is applied in line \ref{algo.state.fp(xt)} in Algorithm \ref{algo} over the graph edges $E$ for $\gamma$ iterations, during which the coordinates in $k_t$ are reset after each step to prevent drift, yielding the propagated view $\hat{X}^{(t)}$.
    
\end{enumerate}

After all $\eta$ iterations are completed, the final representation $\stackrel{*}{X}$ is obtained in line \ref{algo.state.concat} in Algorithm \ref{algo} by concatenating all propagated views column-wise. Formally:

\begin{equation}\label{concat}
\stackrel{*}{X} = \bigoplus_{t=1}^{\eta} \hat{X}^{(t)} \in \mathbb{R}^{|V| \times (d \cdot \eta)}
\end{equation}

where $\oplus$ denotes column-wise concatenation, $|V|$ is the number of nodes, $d$ is the feature dimension of $X$, $\eta$ is the number of generated views, and $\hat{X}^{(t)}$ is the propagated feature matrix from the $t$-th FP view.

Finally, the multi-view representation $\stackrel{*}{X}$ is provided to a downstream GNN, which integrates both the graph structure and the rich node representations to predict node classification $\hat{Y} \in \mathbb{R}^{|V|} $.

\begin{algorithm}
\caption{Multi-View Feature Propagation (MFP) Framework}\label{algo}
\begin{algorithmic}[1]
\State \textbf{Input:} Graph $G = \{ X, E \}$, feature sampling ratio $p$, number of views $\eta$, features to retain $k$, number of propagation iterations $\gamma$.
\State \textbf{Output:} Multi-view propagated representation $\stackrel{*}{X}$ 

\State $\stackrel{\sim}{X} \gets f(X)$ \label{algo.state.sss} \Comment{Apply stochastic sparse sampling (Section~\ref{algo.sss})}
   
\For{$t=1$ to $\eta$}
    \State Sample $k_t \subseteq k$ by $p$ \label{algo.state.get_kt}
    \State Sample $\stackrel{\sim}{X}^{(t)}$ from $\stackrel{\sim}{X}$ by $k_t$ \label{algo.state.get_Xt}
    \State $\hat{X}^{(t)} \gets FP(\stackrel{\sim}{X}^{(t)}, E)$ \label{algo.state.fp(xt)} 
        \Comment{Feature propagation with coordinate reset}
\EndFor
\State $\stackrel{*}{X} \gets \bigoplus_{t=1}^{\eta} \hat{X}^{(t)}$ \label{algo.state.concat} 
    \Comment{Concatenate all propagated views}
\State \Return $\stackrel{*}{X}$
\end{algorithmic}
\end{algorithm}

This framework design, which combines the structural smoothing of FP with the robustness of multi-view learning, is expected to provide two principal benefits:
\begin{enumerate}
    \item \textbf{Privacy Preservation}: By operating on $\stackrel{\sim}{X}$, which exposes only a randomly sampled subset $k$ of features with Gaussian noise injected into the unobserved dimensions, the framework inherently limits the leakage of sensitive information. Unlike conventional FP methods, this stochastic multi-view design ensures that each view contains only a partial and noise-perturbed representation of the original feature space, thereby substantially reducing the risk of reconstructing private attributes.
    \item \textbf{Predictive Performance}: The representation $\stackrel{*}{X}$ is expected to maintain high predictive accuracy relative to the original matrix $X$, as the concatenated embeddings form a rich and expressive feature space that integrates multiple partial perspectives.
\end{enumerate}

In the following sections we test these principles, in addition to evaluating other aspects of the MFP framework’s performance. 


\section{Evaluation Methodology}
\label{Experiments.settings}
We evaluate the MFP framework across a series of experiments designed to examine its node classification performance, privacy-preserving properties, and robustness to different modeling choices. We first assess its effectiveness in node classification tasks by comparing it against several state-of-the-art baselines (Section~\ref{Experiments.performance}). We then conduct dedicated experiments to investigate the degree to which MFP protects sensitive features; in these experiments, we quantify the differences between the transformed feature matrix $\stackrel{*}{X}$ and the original matrix $X$ (Section~\ref{Experiments.privacy}). Finally, we conduct a sensitivity analysis to evaluate the stability of MFP under varying hyperparameter configurations and levels of graph homophily (Section~\ref{Experiments.mfp_sensativity_check}).

\subsection{Datasets}
To evaluate the performance and privacy-preserving capabilities of MFP, we conduct experiments on the three widely used Planetoid benchmark datasets: Cora, Citeseer, and Pubmed \citep{planetoid}. These datasets are standard benchmarks for graph learning algorithms and have also been employed in prior work to investigate privacy issues in graph-structured data \citep{risk24_planetoid}, which makes them particularly suitable for our study. In our setting, we treat the feature matrix $X$ of each dataset as containing sensitive and potentially private attributes whose exposure must be minimized. This aligns with real-world scenarios in which node features often include demographic, behavioral, or textual information that can reveal personal details. Basic statistics for each dataset are summarized in Table \ref{datasets}.

\begin{table}[h]
    \centering
    \caption{\small Statistics of the Planetoid benchmark datasets.}
    {\small
    \begin{tabular}{lccccc}
        \hline\hline
        Dataset & Nodes & Features & Edges & Classes & Homophily \\
        \hline
        Cora     & 2,708  & 1,433 & 10,556  & 7 & 81.0\% \\
        Citeseer &  3,327  & 3,703 & 9,104   & 6 & 73.6\% \\
        Pubmed   &  19,717 & 500   & 88,648  & 3 & 80.2\% \\
        \hline\hline
    \end{tabular}}
    \label{datasets}
    \vspace{-1ex}
\end{table}

To quantify homophily, we adopt the following standard measure: 
\begin{equation}\label{homophily}
\text{Homophily}(G) = \frac{1}{|E|}\sum_{(i,j) \in E} \mathbb{I}_{\{ Y_i = Y_j \} } .
\end{equation}

While the Planetoid datasets provide valuable benchmarks, their homophily levels are fixed (as shown in Table~\ref{datasets}), which limits their suitability for analyzing the effect of varying homophily on MFP. To overcome this limitation, we additionally employ the synthetic MixHop dataset \citep{mixhop}. This dataset consists of 10 graphs with identical node features and label distributions (5,000 nodes across 10 classes) but with edge sets constructed to yield homophily ratios ranging from 0.0 (fully heterophilic) to 0.9 (highly homophilic). Node features are sampled from overlapping two-dimensional distributions specific to each class. This controlled setup enables a systematic study of how different homophily levels influence MFP’s performance.

\subsection{Baselines}

To evaluate the performance of MFP in a node classification task, we compare it with the following baselines: 

\begin{enumerate}
    \item \textbf{Label Propagation (LP)} \citep{label_prop}, which propagates the known labels through graphs based on the similarity between nodes, iteratively updating until convergence. 
    
    \item \textbf{Positional Encoding (PE)} \citep{pos_encode}, which derives node embeddings from the eigenvectors of the graph Laplacian, enabling the model to capture global structural information by encoding each node’s relative position within the graph.
    \item \textbf{Graph Convolutional Network for Missing Features (GCNMF)} \citep{gcnmf}, an end-to-end architecture that addresses feature sparsity by modeling missing values with a Gaussian Mixture Model (GMM).
    \item \textbf{Partial Graph Neural Networks (PaGNN)} \citep{pagnn}, which addresses feature sparsity by introducing partial aggregation functions that use only available neighbor features during message passing, enabling end-to-end learning tasks. 
    \item \textbf{Feature Propagation (FP)} \citep{FP}, our main baseline method, fully described in Section \ref{preliminary.fp}.
    \item \textbf{Random Feature Propagation (RFP)} \citep{RFP}, which propagates randomly sampled features across several trajectories and combines them into a high-dimensional representation vector for each node.
\end{enumerate}

In addition, we include the simple GCN model \citep{gcn} trained on the full feature matrix as a reference baseline representing the \textit{no-privacy, full-information} scenario. Comparing the performance of all sparse-feature methods to this full-feature GCN highlights the trade-off between \textit{privacy preservation and predictive performance}, revealing how effectively each approach maintains accuracy under \textit{extreme feature sparsity and privacy-aware constraints}. It should be noted that MFP, PaGNN, GCNMF, and FP all operate with only 1\% of the original features, with the remaining features randomly removed, whereas PE, LP, and RFP are feature-agnostic. For FP, RFP, and MFP, we employ two GCN layers as the backbone GNN classifier. We use GCN layers rather than alternative GNN architectures (e.g., GAT \citep{gat}), following the setup in \cite{FP}.

To evaluate privacy exposure, we consider two baselines. First, for each dataset, we generate a normally distributed random matrix with the same dimensions as $X$, serving as a reference with zero exposure of sensitive features and allowing us to quantify exposure attributable solely to random noise. Second, we use the output matrix $\hat{X}$ produced by the standard FP method as a baseline for comparison.

\subsection{Metrics}

We adopt the evaluation protocol of \cite{FP} to assess both node classification performance and feature exposure.

\textbf{Classification metrics}. We report accuracy (mean and standard deviation against the ground-truth test labels in $Y$) as a baseline measure. In addition, we emphasize the F1-score as the primary metric, since it balances precision and recall, thereby capturing the trade-off between correctly identifying relevant instances and avoiding false positives.

\textbf{Feature exposure metrics}. To evaluate the extent of private information leakage, we measure the similarity between the original feature matrix $X$ and the propagated output $\stackrel{*}{X}$. First, we compute the root mean square error (RMSE) between each column vector $X_{\cdot c}$ and its $\eta$ propagated variants in $\stackrel{*}{X}$, retaining the minimum RMSE as the closest match. This quantifies the magnitude of the amount of detail preserved from the original features.

Second, to assess directional similarity, we compute the Pearson Correlation Coefficient (PCC) between each feature vector in $X$ and its $\eta$ propagated variants, selecting the highest absolute correlation. PCC highlights whether the propagation preserves structural orientation, which is critical when considering privacy leakage through latent similarity. Together, RMSE and PCC results are summarized using box plots to illustrate the degree of private feature exposure.

\textbf{Cross-representation generalization metrics}. Finally, we evaluate the similarity of propagation output $\hat{X}$ to the original $X$ through model transferability. Let $\Phi^{(X)}$ denote a GNN trained on $X$ (two GCN layers, following \citep{FP}). We test $\Phi^{(X)}$ on $\hat{X}$ and vice versa, yielding predictions $\hat{Y}^{(X, \hat{X})} = \Phi^{(X)}(\hat{X})$ and $\hat{Y}^{(\hat{X}, X)} = \Phi^{(\hat{X})}(X)$. If propagation faithfully reconstructs $X$, cross-domain performance should approach the baselines $\hat{Y}^{(X,X)}$ and $\hat{Y}^{(\hat{X}, \hat{X})}$. We report F1-scores for these cross-generalization experiments, which directly indicate how well propagated features approximate the original space.

\subsection{Experiment Settings}
For the performance evaluation, we chose to adapt the settings presented in \cite{FP}. We conducted 10 runs, each time splitting the graph nodes into training, validation, and test sets. The sets were generated randomly by assigning 20 nodes per class to the training set, 1500 nodes in total to the validation set and the rest to the test set.

For our next experimental scenarios, we conducted 30 runs to calculate variance across runs and assess the stability of our results. In each iteration of our experiments, we removed 99\% of the original features from each dataset to simulate extreme feature sparsity (same as in \cite{FP}). For each single iteration, we randomly split the labels into a training set (80\%) and a test set (20\%). We then used our MFP method to generate a feature matrix based on multi-view propagation, and used a GNN model composed of two GCN layers \citep{gcn} on that matrix for node classification. The GCN was first trained on the training set using Cross Entropy Loss and ADAM optimizer for 100 epochs. The model performance was then evaluated on the test set. 

All of our experiments were executed on a machine equipped with an 8-core Intel i7 processor, 64GB of RAM, and a GeForce RTX 3070 GPU with 8GB of memory.

\subsection{Hyperparameter Settings}
\textbf{Baseline comparisons}. For the performance evaluation against baselines (Section~\ref{Experiments.performance}), we adopted the hyperparameter configurations reported in \cite{FP}. For our proposed MFP model, the number of views was fixed to $\eta = 10$, and the number of propagation iterations was set to $\gamma = 40$, mirroring the FP setup. The stochastic sparse sampling component employed Gaussian noise $\epsilon \sim \mathcal{N}(0,1)$ with a sampling ratio of $p = 0.8$. For RFP, we used $\gamma = 40$, a feature dimension of 16, and $\eta = 10$ trajectories. All other baselines (GCNMF, PaGNN, PE, and LP) were initialized with the hyperparameters specified in \cite{FP}.

\textbf{Privacy and sensitivity analysis}. In the privacy preservation and sensitivity experiments (Sections~\ref{Experiments.privacy} and \ref{Experiments.mfp_sensativity_check}), both FP and MFP were initialized with $\gamma = 40$ as the default setting. To study the effect of propagation depth, we varied $\gamma \in {2, 4, 8, 16, 32, 64, 128}$. To analyze the effect of the number of views, we tested $\eta \in {1, 2, 5, 10, 15, 20}$, where $\eta = 1$ corresponds to the standard FP.

\textbf{Aggregation strategies}. We also experimented with alternative strategies for combining multiple views, such as mean pooling. However, simple concatenation (Algorithm~\ref{algo}, line~\ref{algo.state.concat}) consistently delivered better performance. We therefore adopted concatenation as the default aggregation method and report results using this strategy throughout our experiments.

\section{Results}
\label{results}
In this section, we comprehensively evaluate the proposed MFP framework across three complementary dimensions. We begin by assessing prediction performance under extreme feature sparsity, demonstrating in Section~\ref{Experiments.performance} how MFP maintains strong utility in scenarios where conventional GNNs typically degrade. Next, in Section~\ref{Experiments.privacy}, we investigate the privacy preservation of sensitive features, showing how the multi-view propagation strategy limits the exposure of private attributes while retaining predictive richness. Finally, in Section~\ref{Experiments.mfp_sensativity_check} we present a sensitivity analysis that probes key design factors of MFP, including the effect of graph homophily on model performance, the role of the number of views, and the impact of propagation depth, thereby offering a deeper understanding of the framework’s robustness and design trade-offs.

\subsection{Prediction Performance Under Extreme Feature Sparsity}\label{Experiments.performance}
Table~\ref{results} reports node classification accuracy when 99\% of the original features are randomly removed, simulating an extreme privacy-preserving scenario. We compare MFP against state-of-the-art graph learning methods, including feature-agnostic approaches (PE, LP, RFP) and feature-dependent baselines (PaGNN, GCNMF, FP). In addition, we report the performance of a simple GCN model trained on the full feature matrix, representing the \textit{no-privacy, full-information} setting. This comparison highlights the trade-off between privacy preservation and predictive performance, illustrating how effectively each method maintains accuracy under extreme feature sparsity.


\begin{table}[t]
\centering
\caption{\small Node classification accuracy (mean ± std) of all models under 99\% feature sparsity. Highest accuracy per dataset is in bold. 
As a reference, the rightmost column reports results of GCN trained on the full feature matrices (no privacy).}
\vspace{-0.5ex}
{\footnotesize
\setlength{\tabcolsep}{3pt} 
\renewcommand{\arraystretch}{0.99} 
\begin{tabular}{| l | ccccccc | c |}
\hline\hline
& \multicolumn{7}{c|}{\textbf{99\% Missing features}} & \textbf{Full features} \\
& \multicolumn{7}{c|}{\textit{(Extreme feature sparsity)}} & \textit{(No Privacy)} \\
\hline
Dataset & PaGNN & GCNMF & PE & LP & FP & RFP & \textbf{MFP (ours)} & GCN \\
\hline\hline
Pubmed   & 54.2±0.7 & 39.8±0.2 & 73.7±0.3 & 73.8±0.5 & 74.2±0.5 & 74.8±0.3 & \textbf{76.2}±0.5 & 77.36 \\
Citeseer & 46.0±0.5 & 30.6±1.1 & 65.8±0.3 & 64.6±0.4 & 65.4±0.5 & 65.8±0.2 & \textbf{66.2}±0.2 & 67.48 \\
Cora     & 58.0±0.5 & 34.5±2.0 & 76.3±0.2 & 74.6±0.3 & 78.2±0.3 & 79.3±0.4 & \textbf{80.1}±0.3 & 80.39 \\
\hline\hline
\end{tabular}}
\label{results}
\vspace{-1ex}
\end{table}

As shown in Table \ref{results}, MFP consistently outperforms all baseline methods under extreme feature sparsity. On Cora, MFP achieves the highest accuracy of 80.1\%, compared to 79.3\% with RFP, 78.2\% with FP, and considerably lower values with PaGNN (58.0\%) and GCNMF (34.5\%). On Pubmed, MFP reaches 76.2\%, again surpassing RFP (74.8\%) and FP (74.2\%), and significantly outperforming PaGNN (54.2\%) and GCNMF (39.8\%). Finally, on Citeseer, MFP obtains 66.2\%, higher than RFP (65.8\%) and FP (65.4\%), and well above PaGNN (46.0\%) and GCNMF (30.6\%).

The performance gap underscores two key observations. First, models that rely heavily on original features (PaGNN, GCNMF) degrade severely compared with the \textit{no-privacy} GCN baseline, confirming their limited robustness under extreme sparsity and privacy constraints. Second, feature-agnostic baselines (PE, LP) perform relatively well, but remain inferior to propagation-based methods (FP, RFP, MFP) that explicitly model feature diffusion.

Moreover, when comparing the classification performance of MFP to a standard GCN trained on the complete feature matrix (right column of Table~\ref{results}), we observe that MFP achieves accuracy levels closely matching those of the full-feature model, despite the absence of most input features. This comparison illustrates the fundamental trade-off between \textit{privacy preservation} and \textit{predictive performance}: whereas the full-feature GCN represents the no-privacy upper bound, MFP attains comparable accuracy under an extreme feature-sparsity and privacy-preserving regime. These results demonstrate that MFP effectively retains the predictive capacity of the complete model, even when 99\% of the node features are unavailable.


The consistent improvement of MFP over FP and RFP demonstrates the benefit of multi-view propagation. By generating multiple stochastic propagation instances and aggregating them, MFP enriches the learned feature space and reduces the risk of overfitting to any single propagation trajectory. This mechanism allows the model to retain useful structural signals even when almost all raw features are absent.

\subsection{Privacy Preservation of Sensitive Features}\label{Experiments.privacy}
After establishing the classification performance of MFP, we now evaluate its implications for privacy protection. We benchmark against two baselines. The first is random matrices, which serve as a reference with zero feature exposure, and the second is FP, which represents both the closest methodological baseline and the strongest baseline among the comparison methods in terms of classification accuracy.

\begin{figure}[ht]
    \centering
    \includegraphics[width=0.99\textwidth]{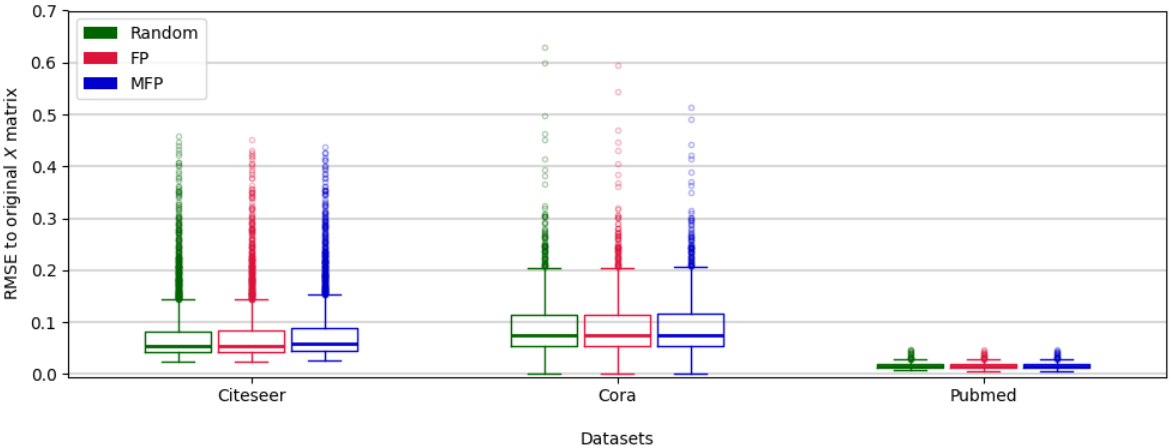}
    \caption{Distance between MFP and FP outputs to the original $X$ features, compared to random noise}
    \label{distance_fig}
\end{figure}

We first quantify the similarity between the output feature matrices and the original feature matrix $X$ using RMSE. Lower RMSE values indicate greater alignment with $X$, and hence higher risk of feature exposure. Figure~\ref{distance_fig} shows the RMSE distributions for Random, FP, and MFP across all datasets.

As illustrated in Figure~\ref{distance_fig}, the results indicate that the majority of feature columns exhibit minimal data leakage. Across all three datasets, both FP and MFP display RMSE distributions that are highly similar to those obtained from randomly sampled data. These findings suggest that the output distances from $X$ generated by both methods are statistically comparable to randomized values, thereby supporting their effectiveness in mitigating information leakage.

In particular, the results reveal that both FP and MFP produce RMSE values largely indistinguishable from those of random noise. This suggests that, at the level of raw magnitude, the propagated features do not reconstruct the original ones and exhibit minimal direct leakage. Importantly, MFP achieves this privacy preservation with superior node classification performance. 

\begin{figure}[ht] \centering 
\includegraphics[width=0.99\textwidth]{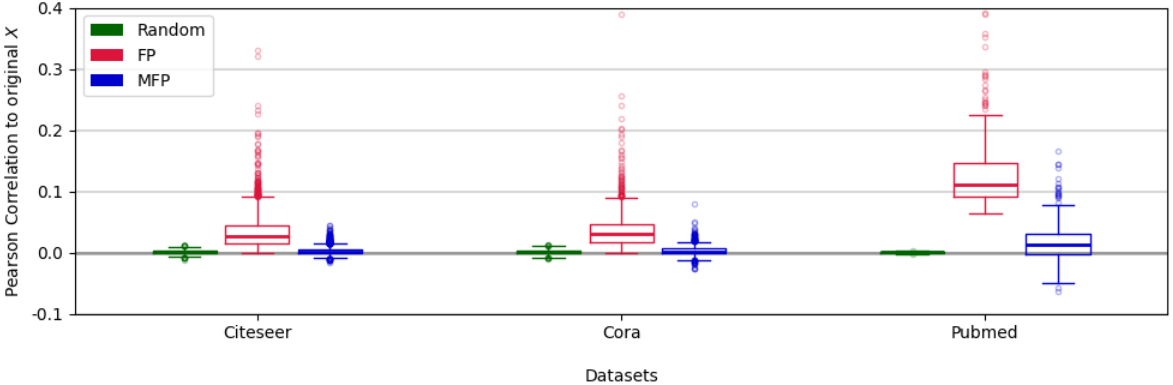} \caption{PCC of $X$ vectors with MFP and FP corresponding output} 
\label{Pearson_fig} 
\end{figure}

Next, we evaluate the correlation between the original feature matrix $X$ and the output matrices (Random, FP, and MFP). In contrast to RMSE, which measures magnitude-based distance, Pearson Correlation Coefficient (PCC) quantifies the degree of linear association, thereby capturing whether output features preserve the directional patterns of the original features. These results are summarized in Figure~\ref{Pearson_fig}.

As shown in Figure~\ref{Pearson_fig}, the output of MFP exhibits very low similarity to the original feature matrix. Specifically, the majority of PCC values are concentrated between –0.1 and 0.1, indicating weak or negligible correlation between $\stackrel{*}{X}$ and $X$. In contrast, FP produces noticeably higher correlation values, reflecting its reliance on a single propagation trajectory, which tends to preserve low-variance signals from the input. By comparison, MFP distributes the retained $k$ features across multiple ($\eta$) propagation views, thereby reducing direct alignment with the original features.

These findings have two important implications. First, they provide further evidence that MFP effectively prevents leakage of structural patterns from sensitive features, since high correlations would otherwise signal partial reconstruction of $X$. Second, they highlight the strength of the multi-view strategy in comparison with FP: by diversifying the propagation process, MFP not only improves predictive performance (Section~\ref{Experiments.performance}) but also attenuates direct similarity to private features, achieving a more robust privacy–utility trade-off.

\begin{figure}[ht]
    \centering
    \includegraphics[width=0.99\textwidth]{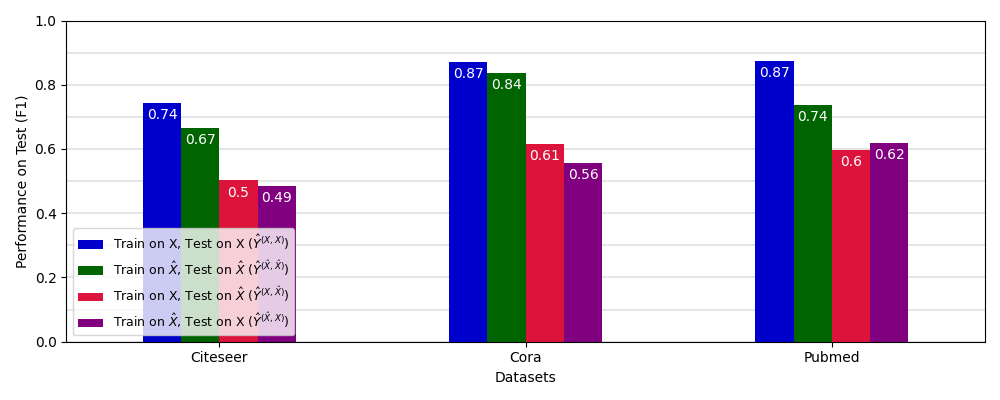}
    \caption{GNN Model F1 performance under each permutation of $X$ and $\hat{X}$, as Train and Test sets}
    \label{model_fig}
\end{figure}

Finally, we analyze the performance of a GNN trained on one representation and evaluated on the other, in order to test whether the propagation process reconstructs the original feature matrix $X$. Figure~\ref{model_fig} presents results across all permutations of training and testing on $X$ and $\hat{X}$.

As shown in Figure~\ref{model_fig}, models trained and evaluated on the same representation perform strongly. For example, on the Cora dataset, $F1(Y,\hat{Y}^{(X,X)}) = 0.87$ and $F1(Y,\hat{Y}^{(\hat{X},\hat{X})}) = 0.84$. The slight gap is expected, since $X$ represents the original feature space while $\hat{X}$ is derived from propagated features. However, when training and testing are performed on different representations, performance drops substantially. On Cora, cross-representation scores fall to $0.61$ for $\hat{Y}^{(X,\hat{X})}$ and $0.56$ for $\hat{Y}^{(\hat{X},X)}$. Similar trends are observed for Citeseer and Pubmed.

These results indicate that the propagated output cannot be regarded as a reconstructed version of $X$. If $\hat{X}$ were a close approximation of $X$, cross-representation evaluations would preserve high performance. Instead, the substantial drop shows that $\hat{X}$ requires entirely different model parameters to be exploited effectively. Thus, $\hat{X}$ should be interpreted as an alternative representation of the missing features rather than as a noisy reconstruction of the original ones.

In summary, the evidence across distance, correlation, and cross-representation analyses shows that MFP does not reconstruct sensitive features but instead generates distinct alternative representations. This multi-view propagation method achieves strong predictive performance while minimizing leakage risk, demonstrating that multi-view propagation offers a practical path toward privacy-preserving graph learning, applicable in scenarios where graph-structured data must be shared, analyzed, or published under strict privacy constraints.

\subsection{Sensitivity Analysis}\label{Experiments.mfp_sensativity_check}
We now present a series of sensitivity analysis experiments (i.e., ablation analysis) for MFP. These experiments examine the effect of (i) the homophily ratio, (ii) the number of propagation iterations ($\gamma$), and (iii) the number of views ($\eta$) on model performance. By systematically varying these factors, we evaluate the robustness of MFP to its key hyperparameters as well as to different structural properties of the underlying graphs.


\subsubsection{Impact of Homophily on Node Classification Performance}
\begin{figure}[ht]
    \centering
    \includegraphics[width=0.99\textwidth]{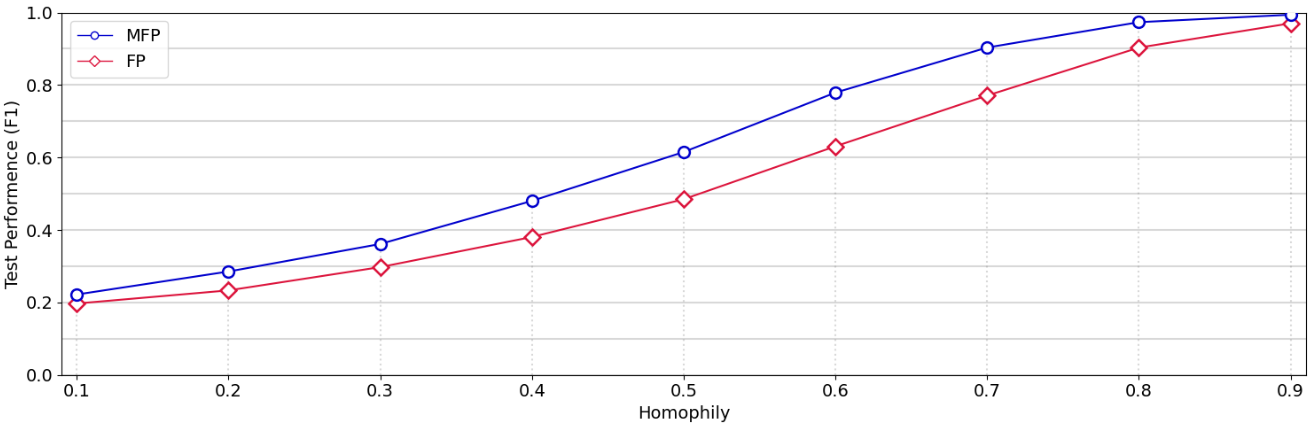}
    \caption{Homophily effect on the performance of FP and MFP}
    \label{homophily_fig}
\end{figure}

As previously discussed, homophily plays a pivotal role in the effectiveness of both FP and GNN-based models \citep{homophily2020, FP}. Figure~\ref{homophily_fig} shows how different levels of homophily affect the node classification performance of FP and of MFP.

As shown in Figure~\ref{homophily_fig}, the performance of both FP and MFP improves monotonically with the homophily ratio. At high homophily levels ($>0.7$), both methods approach high classification performance, consistent with prior findings that homophilous graphs facilitate effective feature propagation \cite{FP}. These results reinforce the central importance of homophily as a structural property that governs the success of propagation-based approaches.

Importantly, MFP consistently outperforms FP across the full spectrum of homophily ratios. This performance gap is most pronounced in low-homophily values ($0.1 \leq h \leq 0.5$), where real-world networks such as citation graphs, recommendation systems, or social platforms often operate \citep{homophily2020}. These findings may suggest that, under low homophily, FP struggles to propagate sufficiently informative signals, while MFP’s multi-view strategy allows it to diversify propagation paths and reduce over-reliance on any single, potentially weak, structural signal. As a result, MFP not only achieves higher performance under favorable conditions but also demonstrates greater robustness when homophily is limited.

These findings highlight a key practical implication: MFP reduces sensitivity to unfavorable graph conditions, making it more reliable for deployment in domains where homophily cannot be guaranteed. For example, in e-commerce or content recommendation graphs, user–item interactions might exhibit low or mixed homophily \citep{low_homophily_bad}; yet accurate predictions are still essential. By mitigating performance degradation in these cases, MFP offers a more dependable solution for organizations seeking to apply graph learning under privacy and sparsity constraints.

\subsubsection{Influence of the Number of Views ($\eta$) on Node Classification Performance}
\begin{figure}[ht]
    \centering
    \includegraphics[width=0.99\textwidth]{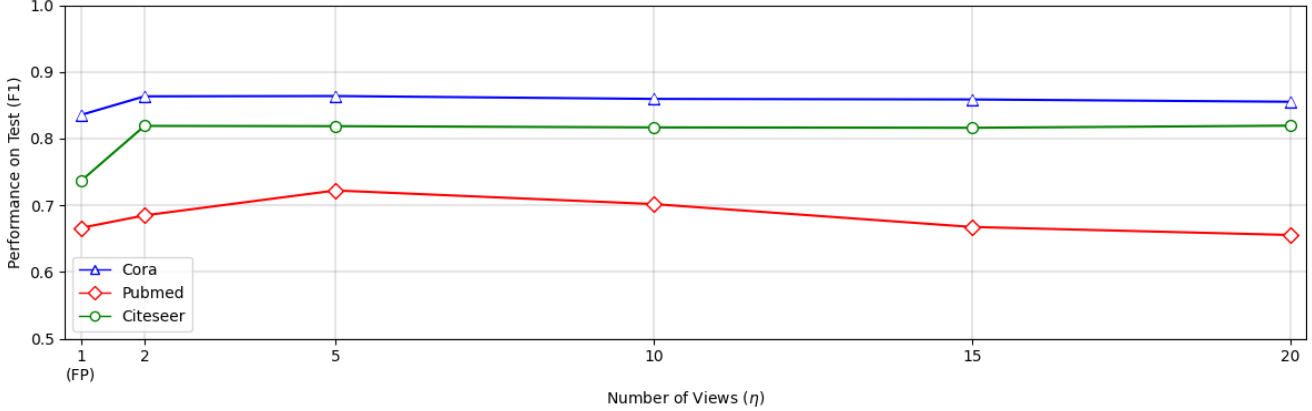}
    \caption{Effect of the number of views ($\eta$) on MFP performance}
    \label{eta_fig}
\end{figure}

We next analyze the influence of the number of views ($\eta$) on node classification performance. The parameter $\eta$ directly affects both the dimensionality of the output representation and the diversity of perspectives used to characterize each node in $G$. Figure~\ref{eta_fig} presents results for varying values of $\eta$ across the Planetoid datasets.

As shown in Figure~\ref{eta_fig}, introducing a small number of views (up to $\eta = 5$) yields a noticeable performance gain relative to the single-view setting ($\eta = 1$), confirming that multi-view propagation enhances node representations by capturing complementary information across subsets of features. Beyond this range, performance saturates for Cora and Pubmed, indicating that once sufficient diversity is achieved, additional views offer limited benefit. For Citeseer, however, a mild decline is observed for larger $\eta$, reflecting a trade-off between the representational gains from view diversity and the increased dimensionality of $\stackrel{*}{X}$, which can introduce redundancy or noise.

Overall, these findings indicate that MFP is robust to the choice of $\eta$: performance gains are achieved for smaller numbers of views (typically with $\eta \leq 10$), and results remain stable across a wide range of values. From a practical standpoint, this robustness is important for deployment. It means that practitioners need not perform extensive tuning of $\eta$ for each dataset, but can rely on moderate default values (e.g., $\eta=10$) to achieve strong performance without incurring unnecessary computational or memory overhead.

\subsubsection{Effect of Propagation Depth ($\gamma$) on Node Classification Performance}

\begin{figure}[ht]
    \centering
    \includegraphics[width=0.99\textwidth]{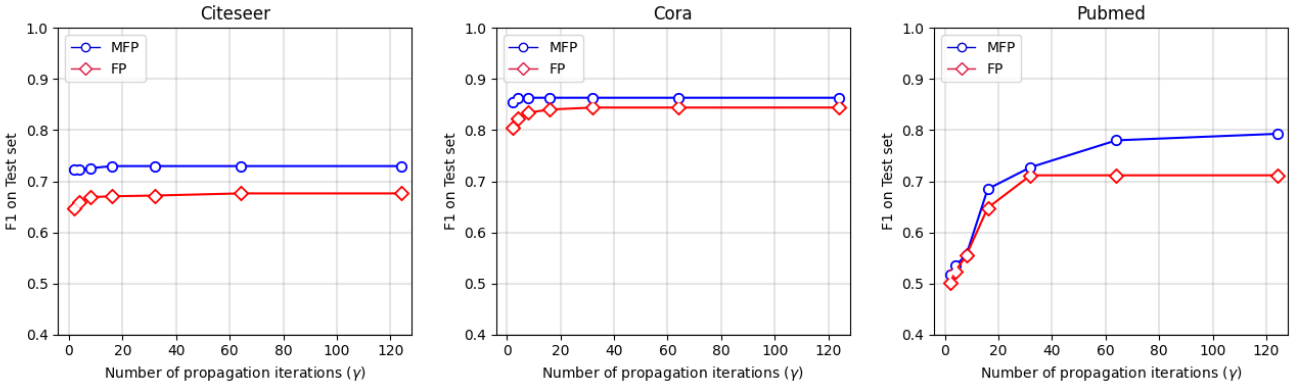}
    \caption{Propagation number ($\gamma$) effect on the performance of FP and MFP}
    \label{gamma_fig}
\end{figure}

We next analyze the influence of the number of propagation iterations ($\gamma$) on the prediction performance. This parameter controls how many times features are diffused across the graph topology. Figure~\ref{gamma_fig} reports the results of MFP and FP across varying values of $\gamma$, illustrating the impact of deeper propagation.

As shown in Figure~\ref{gamma_fig}, increasing $\gamma$ initially yields clear improvements in classification performance. However, the gains quickly diminish and performance converges to a plateau. For Cora and Citeseer, stable performance is achieved after relatively few iterations ($\gamma \leq 16$), while for Pubmed the curve converges more gradually, with performance stabilizing around $\gamma = 64$. This pattern suggests that the optimal number of iterations depends on the structural characteristics of the dataset, such as graph size and connectivity.

Notably, MFP consistently matches or outperforms FP across the entire range of $\gamma$, and importantly, it achieves stability with relatively few propagation steps. This underscores MFP’s robustness to the choice of $\gamma$, which is a desirable property for practical deployment: practitioners do not need to engage in extensive hyperparameter tuning, as small to moderate values of $\gamma$ are sufficient for strong performance.

From a managerial perspective, these results imply that MFP can be deployed efficiently even in resource-constrained environments. Since large values of $\gamma$ provide only marginal gains, organizations can avoid unnecessary computation by choosing default settings (e.g., $\gamma = 40$) without compromising accuracy. This balance between efficiency and robustness enhances the applicability of MFP to real-world graph learning tasks where both scalability and performance are critical.

\section{Conclusion}
In this work, we addressed the dual challenge of achieving robust performance in graph learning under extreme feature sparsity while simultaneously mitigating privacy risks inherent in node-level attributes. We introduced the \textbf{Multi-view Feature Propagation (MFP)} framework, which extends classic FP by leveraging multiple complementary and partially noised views of the feature space. We showed that this framework not only achieves high node classification performance under sparse or noisy data but also reduces the direct exposure of sensitive information.

Specifically, our experimental evaluation across real-world and synthetic datasets demonstrated that MFP consistently outperforms standard propagation and other baseline methods, while offering resilience to privacy risks. The sensitivity analyses further showed that MFP is stable across a wide range of hyperparameters, with consistent results under varying levels of homophily, number of views, and propagation iterations. These findings provide practical guidelines for applying MFP in privacy-aware graph learning tasks, where practitioners can rely on moderate default settings without incurring performance loss.

From a broader perspective, MFP advances the state of privacy-aware graph representation learning by demonstrating that it is possible to balance predictive accuracy, stability, and privacy protection within a unified framework. Beyond the datasets studied here, MFP has the potential to be applied in high-stakes domains such as healthcare, finance, and recommender systems, where feature sparsity and privacy constraints often coexist.

Although MFP shows strong empirical robustness, it has several limitations. First, the current framework assumes random feature sparsity, whereas in practice not all features carry the same level of sensitivity; certain attributes may require stronger protection or differentiated noise injection strategies. Second, while Gaussian noise serves as an effective privacy-enhancing mechanism, integrating formal privacy accounting (e.g., differential privacy) represents an interesting future extension rather than a core requirement of this framework. Finally, while MFP has been validated on diverse benchmark datasets, future work could examine its behavior in larger or dynamic graphs to assess generalizability across application domains.

Future work could extend MFP towards more challenging heterophilous settings, investigate integration with federated or decentralized training paradigms, and explore theoretical bounds on privacy leakage. Taken together, the contributions of this research position MFP as a principled, scalable, and practically deployable solution for modern graph learning applications under privacy constraints.


\bibliographystyle{sn-basic}
\bibliography{sn-bibliography}


\begin{appendices}

\section{Legend of Notions}\label{secA1}

For clarity and ease of reference, we provide a comprehensive table summarizing all the notations used throughout this paper. Table \ref{legend} includes the symbols, their definitions, and descriptions of their roles within our paper. It is intended to help readers quickly understand the mathematical and conceptual terms employed in our work.

\begin{table}[h]
    \centering
    \caption{\small Legend of notions}
    {\small
    \begin{tabular}{l ll}
        \hline
        \hline
        Notion & \parbox[c]{10cm}{description} &   \\
        \hline
        \hline
        $V$ & \parbox[t]{6cm}{Nodes set in the graph} &
        \\
        \\
        \hline
        $i,j$ & \parbox[t]{6cm}{Nodes indices} &
        \\
        \\
        \hline
        $E$ & \parbox[t]{6cm}{Edges set in the graph} &
        \\
        \\
        \hline
        $G$ & \parbox[t]{6cm}{Graph object} & 
        \\
        \\
        \hline
        $X$ & \parbox[t]{6cm}{Nodes features matrix} & 
        \\
        \\
        \hline
        $Y$ & \parbox[t]{6cm}{Node classification vector} & 
        \\
        \\
        \hline
        $\hat{Y}$ & \parbox[t]{6cm}{Predicted Node classification vector} & 
        \\
        \\
        \hline
        $d$ & \parbox[t]{6cm}{Feature dimension of $X$} &
        \\
        \\
        \hline
        $k$ & \parbox[t]{6cm}{Set of retained feature values} &
        \\
        \\
        \hline
        $p$ & \parbox[t]{6cm}{Feature retention ratio} &
        \\
        \\
        \hline
        $c$ & \parbox[t]{6cm}{Column index} &
        \\
        \\
        \hline
        $\hat{X}$ & \parbox[t]{6cm}{Feature Propagation output matrix} & 
        \\
        \\
        \hline
        $f( \cdot)$ & \parbox[t]{6cm}{Stochastic Sparse Sampling function} &
        \\
        \\
        \hline
        $\epsilon_{ic}$ & \parbox[t]{10cm}{Gaussian Noise that sampled from normal distribution} &
        \\
        \\
        \hline
        $\stackrel{\sim}{X}$ & \parbox[t]{10cm}{Stochastic Sparse Sampling output matrix (Section \ref{algo.sss})} & 
        \\
        \\
        \hline
        $\gamma$ & \parbox[t]{6cm}{Number of propagation iterations} & 
        \\
        \\
        \hline
        $\iota$ & \parbox[t]{6cm}{Index of single propagation iteration} & 
        \\
        \\
        \hline
        $\eta$ & \parbox[t]{6cm}{Number of views for MFP algorithm} & 
        \\
        \\
        \hline
        $t$ & \parbox[t]{6cm}{Single view index} & 
        \\
        \\
        \hline
        $\stackrel{*}{X}$ & \parbox[t]{6cm}{Nodes features matrix after MFP} & 
        \\
        \\
        \hline
        $\Phi$ & \parbox[t]{6cm}{GNN classification model} & 
        \\
        \\
        \hline
        $\hat{Y}^{(A,B)}$ & \parbox[t]{10cm}{Node classification vector that created using a model train on representation $A$ and applied on representation $B$} & 
        \\
        \\
        \hline
        \hline
    \end{tabular}}
    \label{legend}
     \vspace{-1ex}
\end{table}




\end{appendices}


\end{document}